\documentclass{article}

\usepackage{microtype}
\usepackage{graphicx}
\usepackage{booktabs} 
\usepackage{pdfpages}

\usepackage{hyperref}

\usepackage[accepted]{icml2023}

\usepackage{amsmath}
\usepackage{amssymb}
\usepackage{mathtools}
\usepackage{amsthm}
\usepackage{subcaption}

\usepackage[capitalize,noabbrev]{cleveref}

\theoremstyle{plain}

\theoremstyle{definition}

\theoremstyle{remark}

\usepackage[textsize=tiny]{todonotes}

\icmltitlerunning{On the Geometry of User Preferences in Large-Scale Collaborative Filtering}

\begin{document}

\twocolumn[
\icmltitle{Latent Geometry of Taste: Scalable Low-Rank Matrix Factorization for Recommender Systems}




\begin{icmlauthorlist}
\icmlauthor{Joshua Salako}{aims}
\end{icmlauthorlist}

\icmlaffiliation{aims}{African Institute for Mathematical Sciences (AIMS) South Africa, 6 Melrose Road, Muizenberg 7975, Cape Town, South Africa}
\icmlcorrespondingauthor{Joshua Salako}{jsalako@aims.ac.za}

\icmlkeywords{collaborative filtering, matrix factorization, recommendation systems, alternating least squares,
latent preference geometry}

\vskip 0.3in
]



\printAffiliationsAndNotice{\icmlEqualContribution} 

\begin{abstract}
Scalability and data sparsity remain critical bottlenecks for collaborative filtering on massive interaction datasets. This work investigates the latent geometry of user preferences using the MovieLens 32M dataset, implementing a high-performance, parallelized Alternating Least Squares (ALS) framework. Through extensive hyperparameter optimization, we demonstrate that constrained low-rank models significantly outperform higher dimensional counterparts in generalization, achieving an optimal balance between Root Mean Square Error (RMSE) and ranking precision. We visualize the learned embedding space to reveal the unsupervised emergence of semantic genre clusters, confirming that the model captures deep structural relationships solely from interaction data. Finally, we validate the system's practical utility in a cold-start scenario, introducing a tunable scoring parameter to manage the trade-off between popularity bias and personalized affinity effectively. The codebase for this research can be found here: \emph{https://github.com/joshsalako/recommender.git}
\end{abstract}

\section{Introduction}
\label{intro}

In the era of information overload, recommender systems have become the fundamental engine driving user engagement across digital platforms, from e-commerce to streaming services. The core challenge lies in effectively filtering the "long tail" of available content to match user preferences, a problem exacerbated by the sparsity of user-item interaction matrices. While deep learning approaches have recently gained traction, Latent Factor Models, specifically Matrix Factorization (MF), remain the cornerstone of collaborative filtering due to their scalability, interpretability, and predictive power \citep{koren2009matrix}.

This work explores the latent structures governing user preferences using the MovieLens 32M dataset, which contains over 32 million ratings. This dataset presents significant computational challenges that render naive implementations infeasible. We implement a highly optimized Alternating Least Squares (ALS) framework, accelerated via Just-In-Time (JIT) compilation, to decompose the interaction matrix into lower-dimensional user and item embeddings. Our contributions are threefold. First, we perform an extensive Exploratory Data Analysis (EDA) on the MovieLens 32M dataset. Second, we implement and compare a Bias-only model against a full MF model, analyzing the trade-offs between RMSE and ranking metrics. Finally, we visualize the geometry of the learned latent space, demonstrating how the model naturally recovers semantic clusters without supervision, and analyze the probabilistic nature of these embeddings through the lens of joint density functions. 

The remainder of this work is structured as follows: Section \ref{dataset} provides a detailed characterization of the MovieLens 32M dataset, including exploratory data analysis and the sparse matrix representation used for efficient processing. Section \ref{als} formulates the ALS algorithm. Finally, Section \ref{result} presents the experimental results, including hyperparameter optimization, visualization of the learned latent geometry, and a qualitative evaluation of recommendation performance in cold-start scenarios.

\medskip
\paragraph{Related Works}

Collaborative filtering strategies generally fall into two categories: neighborhood methods and latent factor models. Early approaches relied heavily on neighborhood methods, which compute relationships between items or users directly \citep{sarwar2001item}. While intuitive, these methods suffer from scalability issues as the dataset size increases. The landscape of recommender systems shifted significantly during the Netflix Prize competition, which demonstrated the superiority of MF techniques. \cite{koren2009matrix} formalized the decomposition of the rating matrix into user and item vectors, allowing for the incorporation of biases and temporal dynamics. Their work highlighted that while Stochastic Gradient Descent (SGD) is easy to implement, ALS is often more favorable for parallelization and handling implicit feedback datasets. Industrial applications require models that scale to millions of users. \cite{koenigstein2012xbox} detailed the Xbox Recommender System, emphasizing that real-world systems must move beyond simple error minimization to optimize for utility and ranking. They utilized a Bayesian framework to handle uncertainty in user profiling. Similarly, Probabilistic MF \citep{salakhutdinov2007probabilistic} frames the recommendation problem as a probabilistic graphical model with Gaussian priors, a concept we revisit in our analysis of the joint density of latent factors.

\section{MovieLens 32M Dataset}
\label{dataset}

We utilize the MovieLens 32M dataset, a large-scale benchmark for collaborative filtering research. The dataset comprises approximately 32 million ratings applied to over 80,000 movies by more than 200,000 users. The ratings follow a 5-star scale with half-star increments.

\subsection{Exploratory Data Analysis}

\begin{figure*}[tb!]
    \centering
    \includegraphics[width=0.8\linewidth]{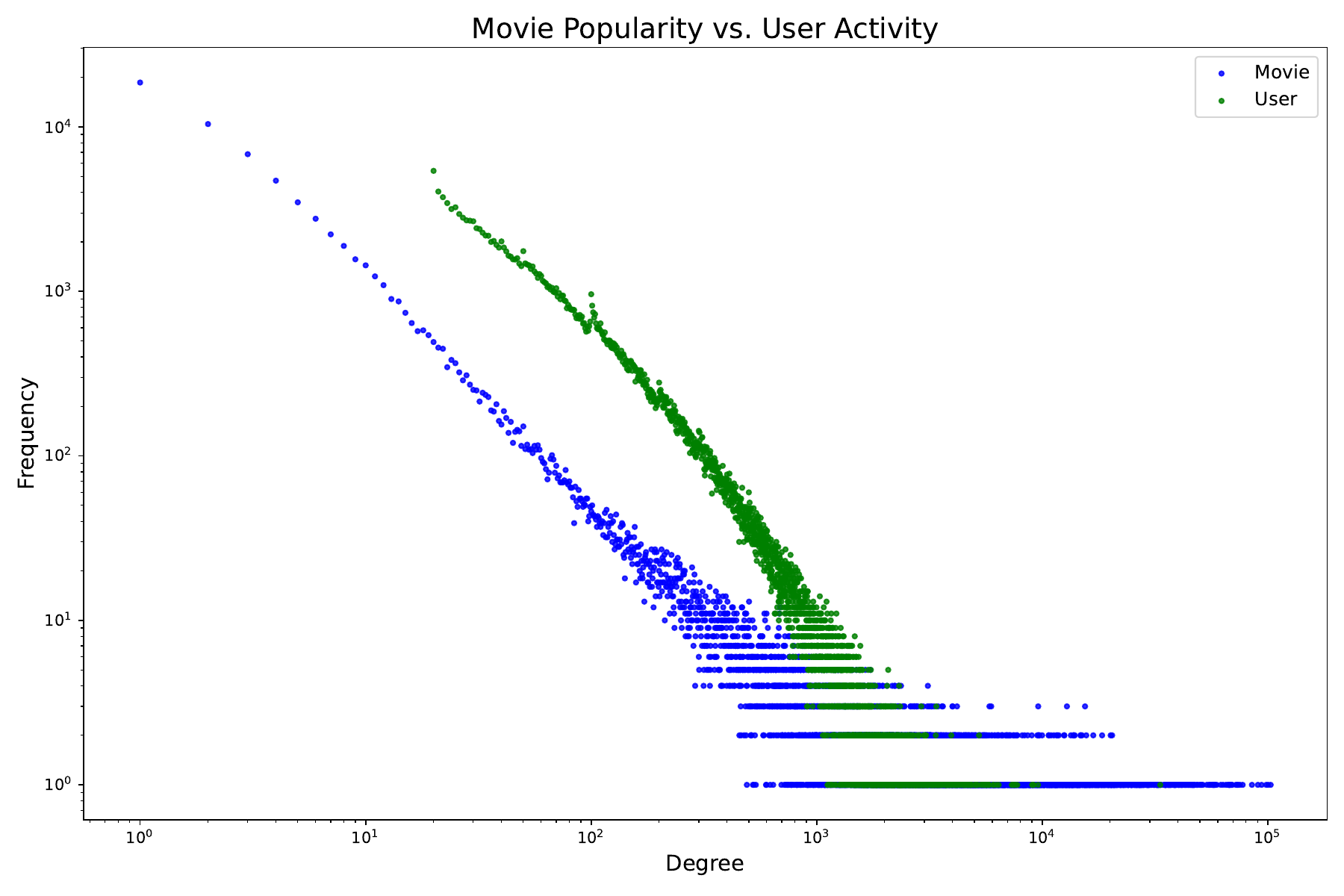}
    \caption{Power law distribution across MovieLens 32M dataset}
    \label{fig:degree_dist}
\end{figure*}

To understand the underlying structure of user preferences, we analyze the distribution of interactions across users and items. As illustrated in Figure~\ref{fig:degree_dist}, the interaction frequency for both users and movies follows a power-law distribution. A small subset of "blockbuster" movies accumulates the majority of ratings, while the vast majority of items reside in the long tail with sparse interactions. Similarly, a small fraction of power users contribute a disproportionate number of ratings.

\begin{figure*}[h!]
    \centering
    \begin{subfigure}[b]{0.49\textwidth}
        \centering
        \includegraphics[width=\linewidth]{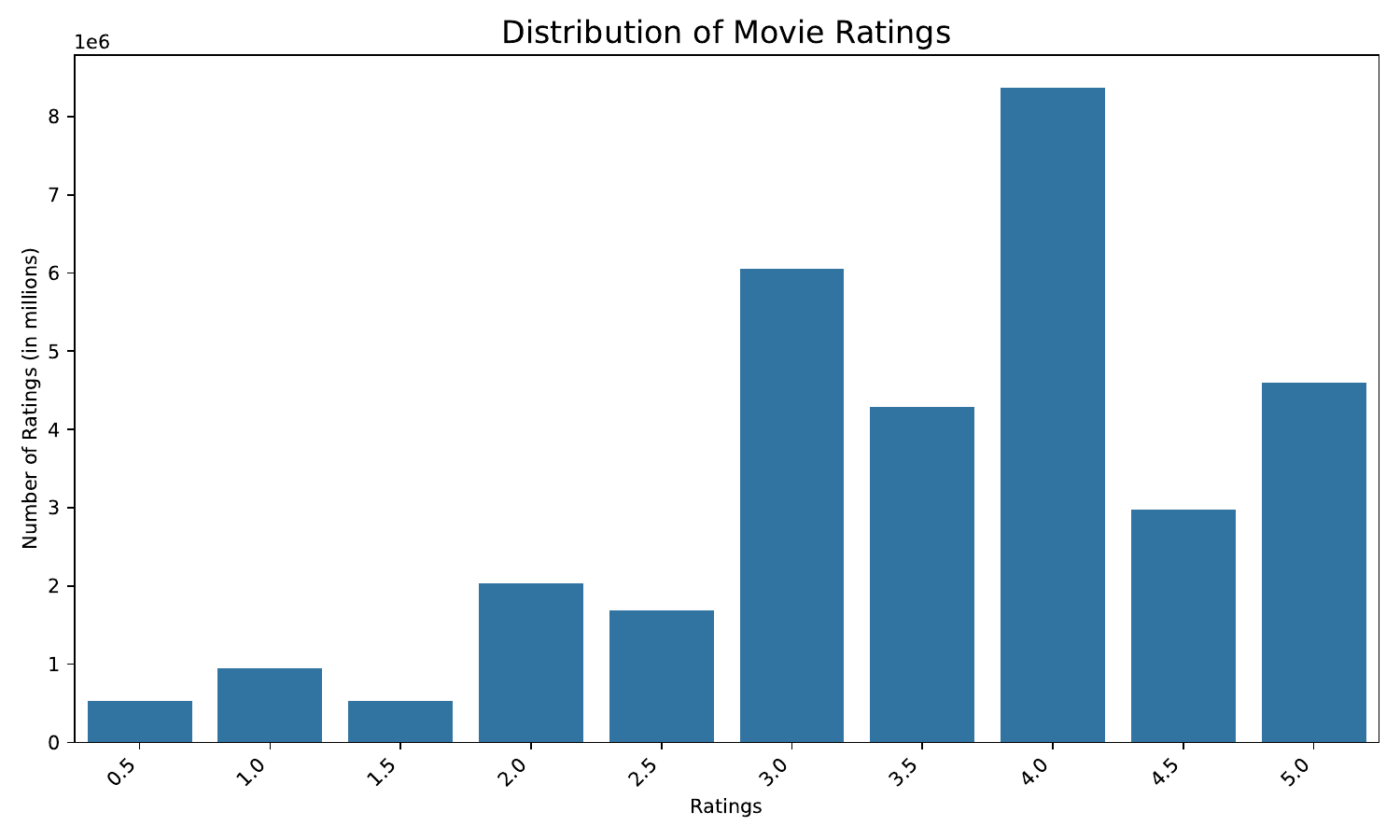}
        \caption{Distribution of rating values.}
        \label{fig:rating_dist}
    \end{subfigure}
    \hfill
    \begin{subfigure}[b]{0.49\textwidth}
        \centering
        \includegraphics[width=\linewidth]{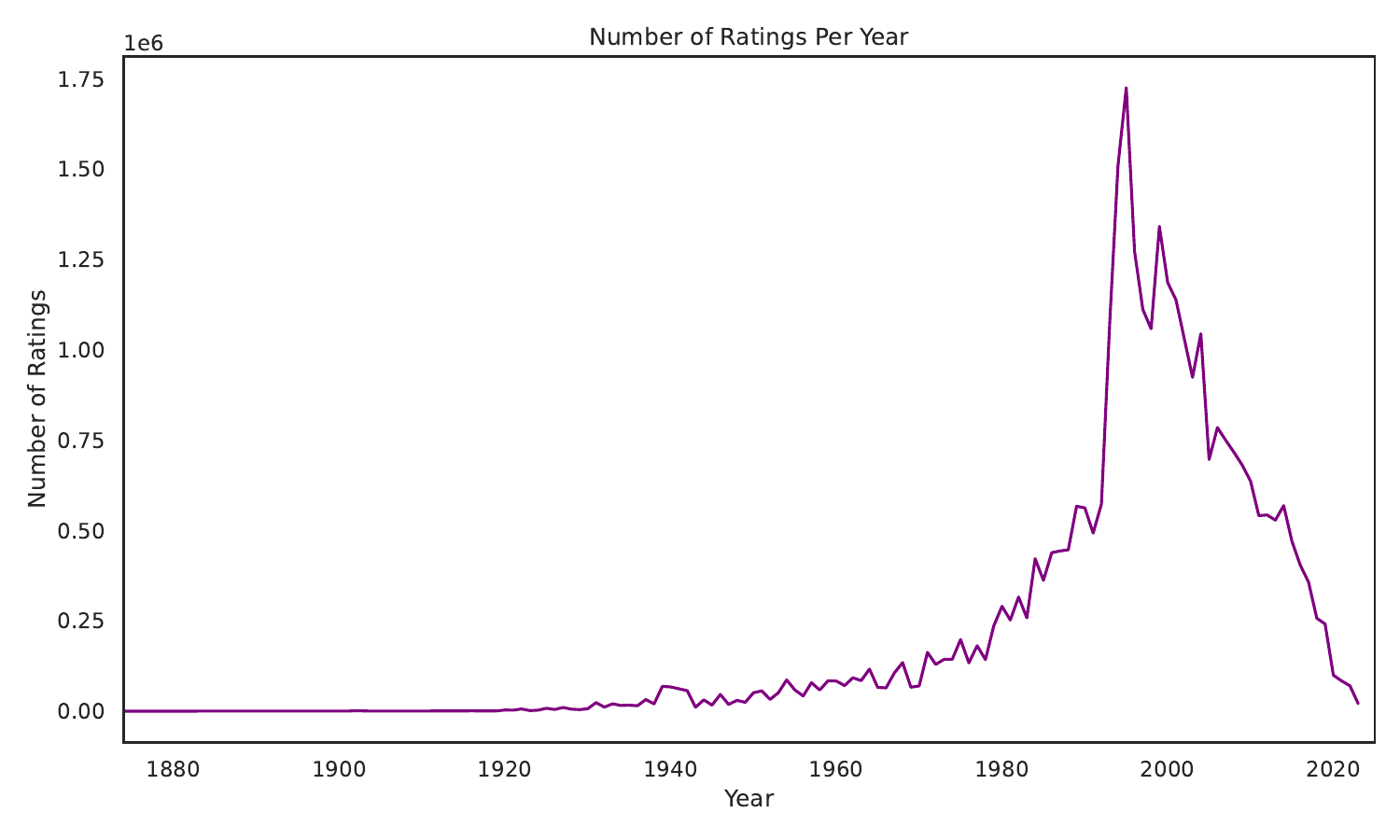}
        \caption{Distribution of movie rating values.}
        \label{fig:ratings_trends_year}
    \end{subfigure}
    
    \vspace{0.5cm}
    
    \begin{subfigure}[b]{0.49\textwidth}
        \centering
        \includegraphics[width=\linewidth]{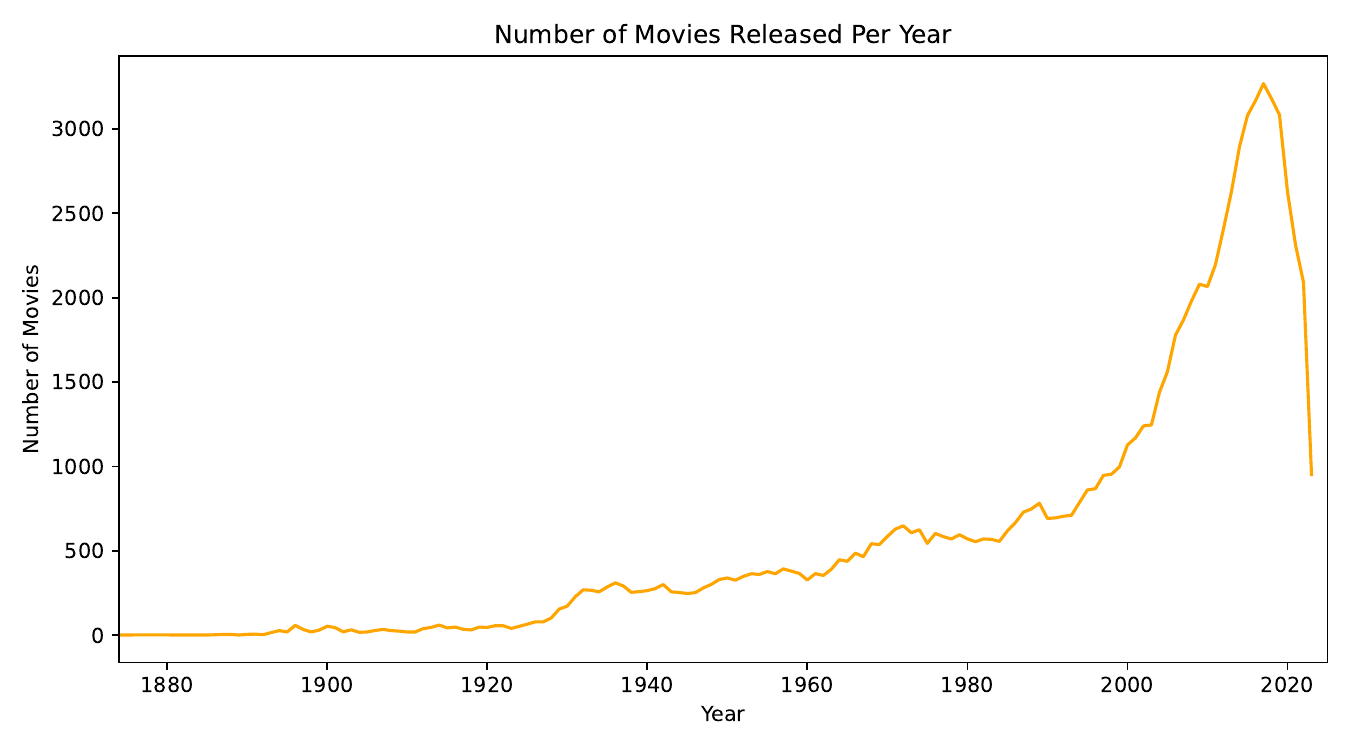}
        \caption{Number of movies released per year.}
        \label{fig:movie_count}
    \end{subfigure}
    \hfill
    \begin{subfigure}[b]{0.49\textwidth}
        \centering
        \includegraphics[width=\linewidth]{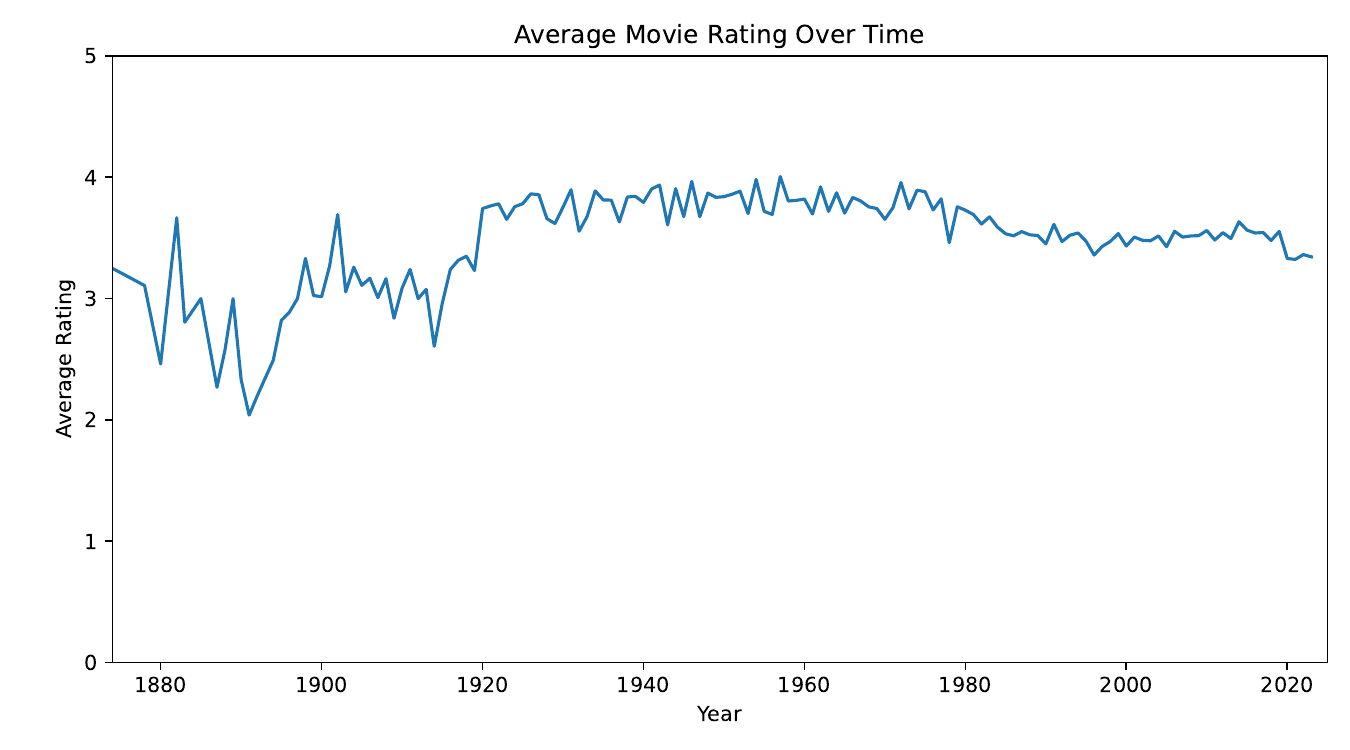}
        \caption{Average movie rating by release year.}
        \label{fig:time_trends}
    \end{subfigure}
    
    
    \begin{subfigure}[b]{0.49\textwidth}
        \centering
        \includegraphics[width=\linewidth]{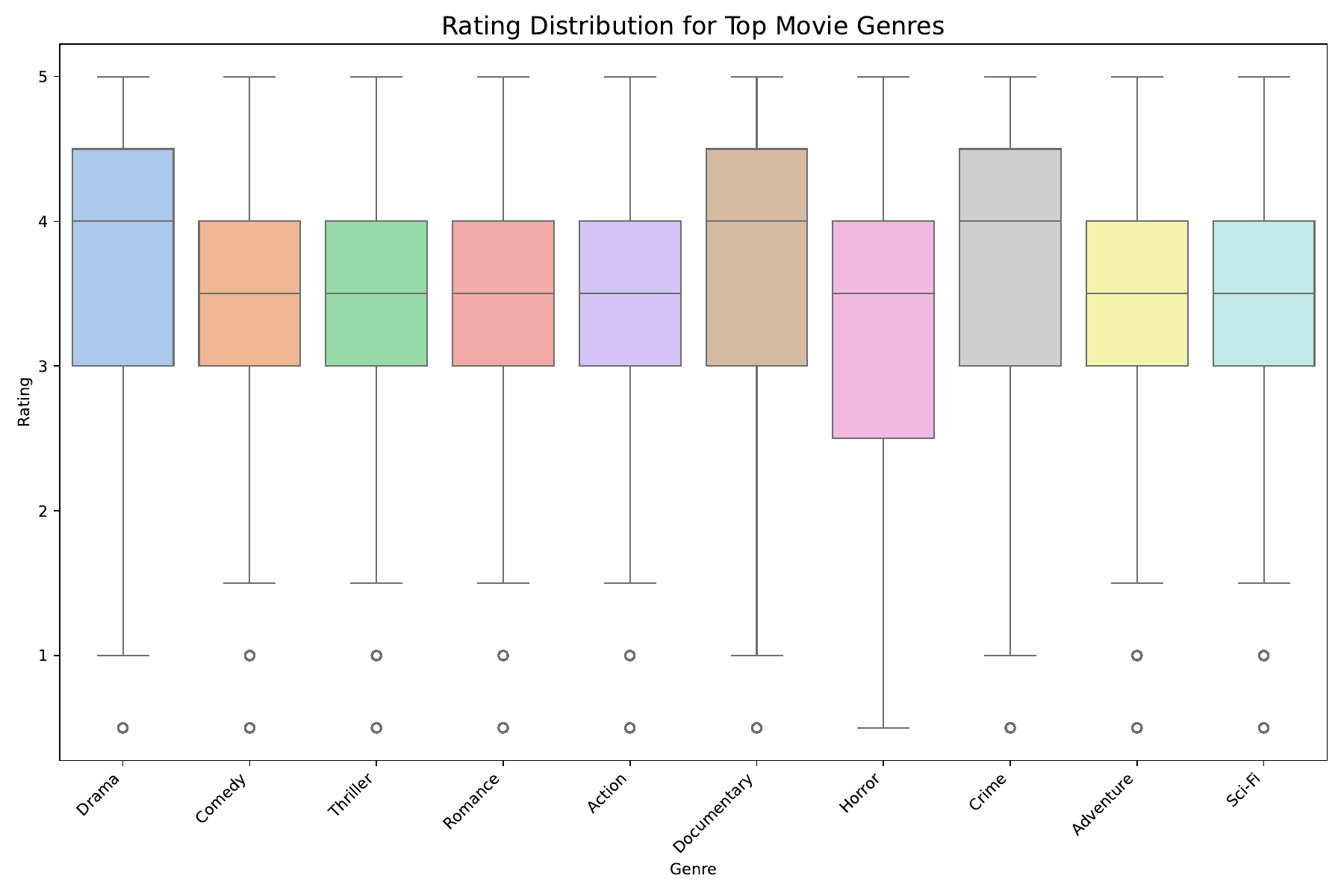}
        \caption{Average ratings for top movie genres.}
        \label{fig:genre_trends}
    \end{subfigure}
    \hfill
        \begin{subfigure}[b]{0.4\textwidth}
        \centering
        \includegraphics[width=\linewidth]{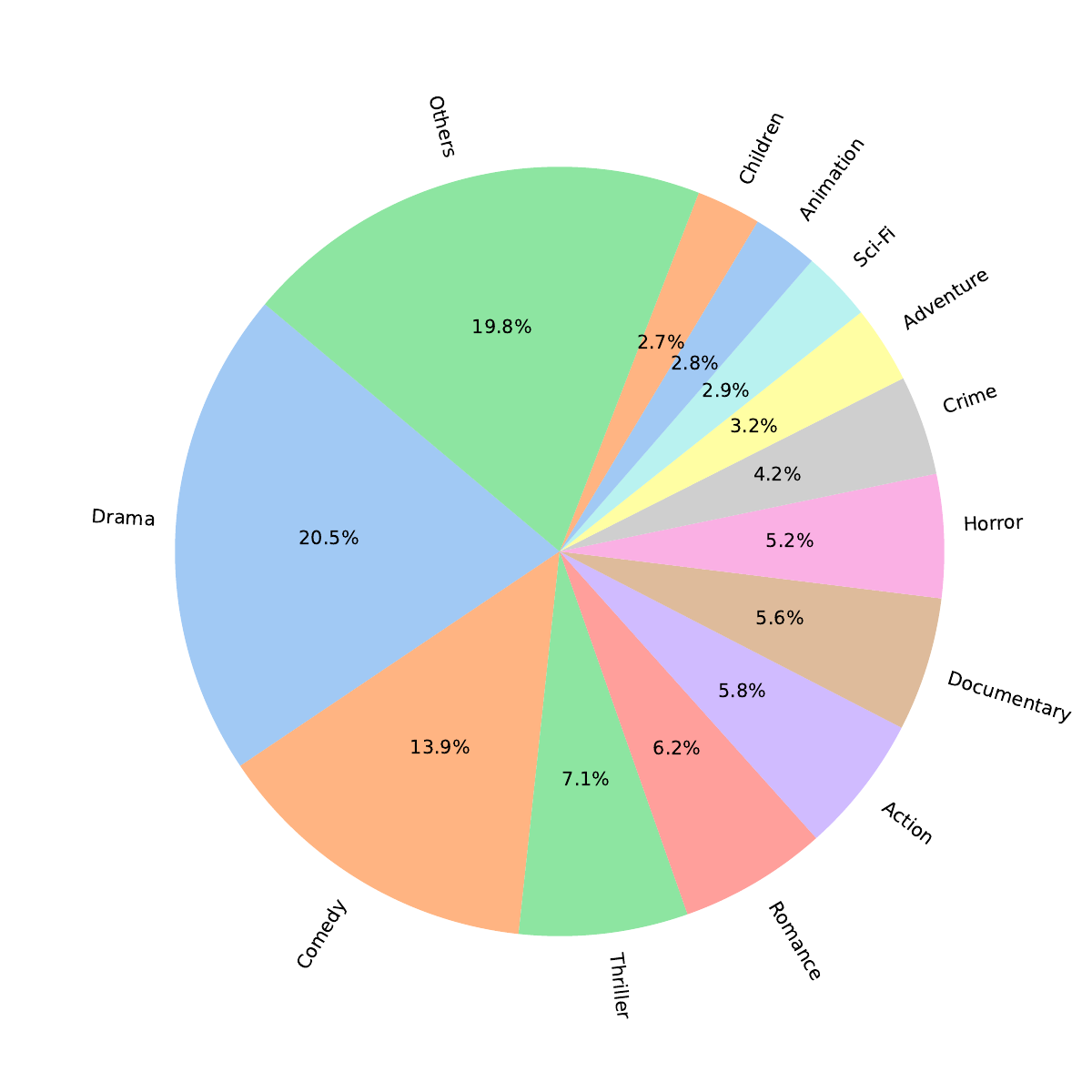}
        \caption{Percentage distribution of top movie genres.}
        \label{fig:genres}
    \end{subfigure}
    
    \caption{Visualizations of MovieLens 32M dataset.}
    \label{fig:all_visualizations}
\end{figure*}

We further examine the distribution of explicit feedback in the MovieLens 32M dataset. As illustrated in Figure~\ref{fig:rating_dist}, the ratings exhibit a left-skewed distribution, with a global mean of approximately 3.53. This pattern reflects a positivity bias, whereby users tend to assign ratings primarily to movies they appreciated. Figures~\ref{fig:movie_count} and~\ref{fig:ratings_trends_year} reveal exponential growth in movie releases and user ratings, respectively, across decades, followed by a sharp decline in recent years, potentially attributable to dataset collection artifacts or shifts in user behavior. Despite these trends, Figure~\ref{fig:time_trends} indicates that average ratings have remained relatively stable over time, fluctuating modestly around the global mean.

\begin{figure}[ht!]
    \centering
    \includegraphics[width=\linewidth]{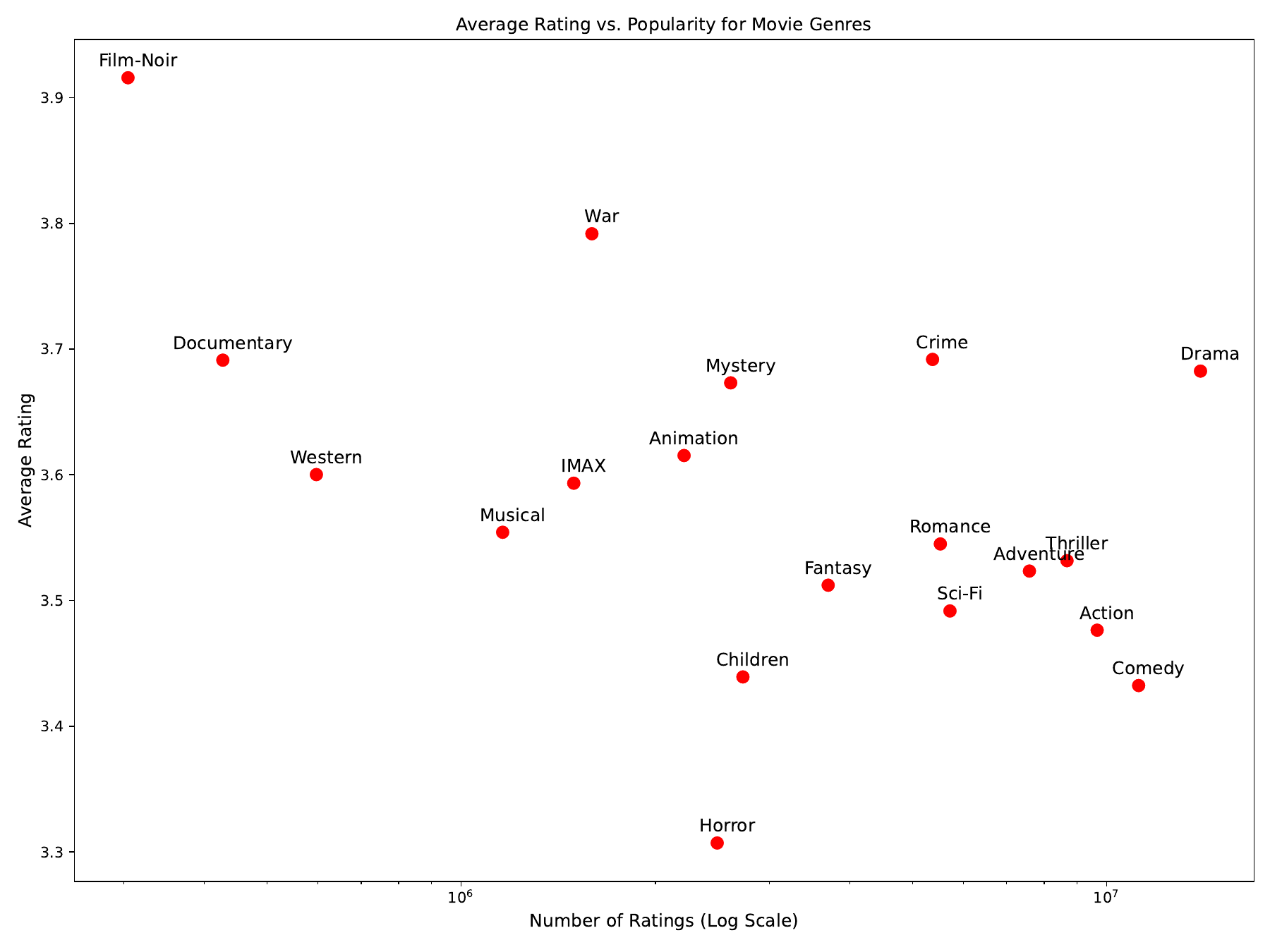}
    \caption{Average rating versus popularity for movie genres}
    \label{fig:genre_pop}
\end{figure}

\begin{figure}[ht!]
    \centering
    \includegraphics[width=\linewidth]{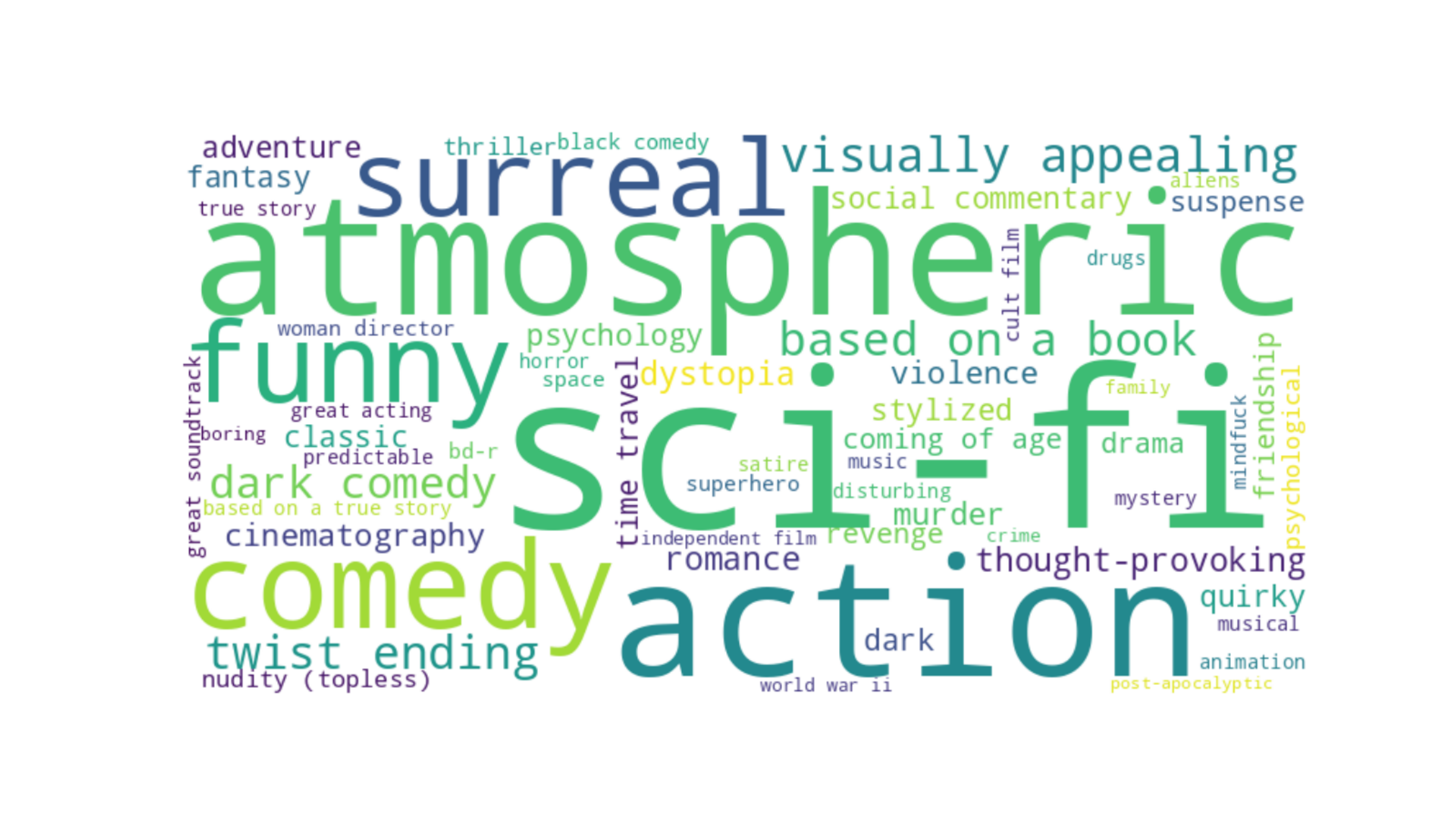}
    \caption{Visualization of user tags across MovieLens 32M dataset.}
    \label{fig:tags}
\end{figure}

To explore semantic patterns, we assess genre related characteristics. Figure~\ref{fig:genre_trends} depicts the average ratings across top genres, revealing variations such as higher means for Drama and Documentary compared to Horror. Figure~\ref{fig:genre_pop} extends this analysis by plotting average ratings against genre popularity, highlighting that niche genres like Film-Noir achieve high ratings despite lower popularity, while mainstream genres like Comedy show moderate ratings with high engagement. Figure~\ref{fig:genres} underscores genre imbalance, with Drama (20.5\%) and Comedy (13.9\%) dominating the corpus, followed by Thriller (7.1\%), Romance (6.2\%), and others; this skew challenges models to capture nuanced features for underrepresented categories without bias toward dominant ones. Finally, Figure~\ref{fig:tags} presents a word cloud of user-assigned tags, emphasizing prevalent themes such as atmospheric, sci-fi, comedy, and action, which provide additional semantic context for movie representations.

\subsection{Data Preprocessing}

To facilitate matrix operations, we implemented an indexing layer that maps raw identifiers to contiguous integers $u \in [0, N_{users}-1]$ and $i \in [0, M_{movies}-1]$. We employed a stratified train-test split strategy to preserve the user activity distribution. For every user, we randomly sample 80\% of their ratings for training and hold out the remaining 20\% for testing. This ensures that all users in the test set have a representation in the training phase, mitigating the user cold-start problem during evaluation.

\subsubsection{Sparse Matrix Representation}

Given the large dimensions of the dataset, a dense matrix representation is computationally intractable. To address this, we implement a custom Compressed Sparse Row (CSR) format optimized for parallel processing. We structure the data using three flat arrays:
\begin{itemize}
    \item \textbf{Values:} An array of rating values $r_{ui}$.
    \item \textbf{Indices:} An array of column indices, for each corresponding rating value, this array stores the identifier of the item that was rated.
    \item \textbf{Offsets:} An array of length $N+1$ is used to delineate the data for each user. For a given user $u$, the contiguous block of entries associated with them in the Values and Indices arrays is defined by the range starting at the $u$-th element and ending just before the $(u+1)$-th element of this Offsets array.
\end{itemize}


\section{Alternating Least Squares}
\label{als}

To uncover the latent structure of the user-item interaction matrix, we employ MF, where the fundamental goal is to approximate the observed rating matrix, $R$, as the product of two lower-dimensional matrices, $U$ and $V$, which represent the user and item, respectively. While SGD is a popular optimization method for this task, we utilize ALS \citep{koren2009matrix}. The primary advantage of ALS in our high-dimensional context is its amenability to parallelization. The objective function for MF is non-convex when optimizing for $U$ and $V$ simultaneously. However, if we fix the user factors $U$, the problem becomes quadratic with respect to $V$, and vice versa. This allows us to solve for the optimal parameters analytically using Ridge Regression in an iterative fashion \citep{zhou2008large}.

\subsection{Bias-only Model}

Before modeling specific user preferences, it is crucial to account for systematic tendencies in the data, known as biases. Certain users tend to rate conservatively, while others are generous. Similarly, some movies enjoy universal acclaim while others are generally disliked, independent of genre.

We define the predicted rating $\hat{r}_{ui}$ in the bias-only model as:

\begin{equation}
    \hat{r}_{ui} = \mu + b_u + b_i
    \label{eq:bias_pred}
\end{equation}

where $\mu$ is the global average rating, $b_u$ is the user bias, and $b_i$ is the item bias. We aim to minimize the regularized squared error. In our implementation, we apply the regularization parameter $\tau$ to the magnitude of the parameters and a weighting parameter $\lambda$ to the sum of squared errors:

\begin{equation}
L = \lambda \sum_{(u,i)\in\mathcal{K}} \bigl(r_{ui} - \hat{r}_{ui}\bigr)^2
    + \tau \Bigl( \sum_u b_u^2 + \sum_i b_i^2 \Bigr)
\label{eq:bias_loss}
\end{equation}

By taking the derivative of Eq. \ref{eq:bias_loss} with respect to $b_u$ and setting it to zero, we derive the closed-form update rule for user biases:

\begin{equation}
b_u = \frac{\lambda \sum_{i \in R(u)} (r_{ui} - \mu - b_i)}{\tau + \lambda \, |R(u)|}
\label{eq:bias_update}
\end{equation}

where $R(u)$ is the set of items rated by user $u$. A symmetric update rule is applied to item biases $b_i$. We iterate between updating all user biases and all item biases until convergence.

\subsection{Bias + Trait Vector Model}

To capture personal taste, we extend the model by introducing latent trait vectors. Each user is associated with a vector $\mathbf{u}_u \in \mathbb{R}^k$ and each item with a vector $\mathbf{v}_i \in \mathbb{R}^k$, where $k$ is the number of latent dimensions. The prediction rule becomes the inner product of these vectors, adjusted by the biases:

\begin{equation}
    \hat{r}_{ui} = \mu + b_u + b_i + \mathbf{u}_u^T \mathbf{v}_i
    \label{eq:full_pred}
\end{equation}

The objective function is expanded to regularize the latent vectors:

\begin{equation}
L = \lambda \sum_{(u,i) \in \mathcal{K}} (r_{ui} - \hat{r}_{ui})^2 + \tau \Bigl( \|b\|^2 + \sum_u \|\mathbf{u}_u\|^2 + \sum_i \|\mathbf{v}_i\|^2 \Bigr)
\label{eq:full_loss}
\end{equation}

In our implementation, we perform a coordinate descent. When updating user $u$, we treat item vectors $\mathbf{v}_i$ and item biases $b_i$ as constants. The update for the user latent vector $\mathbf{u}_u$ is the solution to a regularized linear least squares problem:

\begin{equation}
\mathbf{u}_u = \bigl( \lambda \! \sum_{i \in R(u)} \mathbf{v}_i \mathbf{v}_i^T + \tau \mathbf{I} \bigr)^{\!-1}
               \Bigl( \lambda \! \sum_{i \in R(u)} \mathbf{v}_i (r_{ui} - \mu - b_u - b_i) \Bigr)
\label{eq:als_update}
\end{equation}

where $\mathbf{I}$ is a $k \times k$ identity matrix. We leverage the dual sparse matrix representations described in Section \ref{dataset} to perform these updates efficiently. 


\subsection{Model Training}

We implement the training pipeline using a custom, highly optimized framework designed to handle the scale of the MovieLens 32M dataset. The core ALS update rules are JIT compiled using \texttt{Numba}. The training process alternates between two phases:
\begin{enumerate}
    \item \textbf{User Step:} We iterate through the user-centric CSR structure. For each user $u$, we solve the system of linear equations, Eq. \ref{eq:als_update}, to update $\mathbf{u}_u$ and $b_u$ while keeping all item parameters fixed.
    \item \textbf{Item Step:} We iterate through the item-centric CSR structure, updating $\mathbf{v}_i$ and $b_i$ while keeping user parameters fixed.
\end{enumerate}

Latent vectors are initialized from a normal distribution $\mathcal{N}(0, \frac{1}{\sqrt{k}})$, while biases are initialized to zero. We train the model for 20 epochs while monitoring the loss function and RMSE after every iteration to ensure convergence.

\subsection{Model Evaluation}
To comprehensively assess the quality of the recommender system, we employ two distinct categories of metrics: predictive accuracy and ranking quality.

\paragraph{Root Mean Squared Error (RMSE)}
This measures the average magnitude of the error between the predicted rating $\hat{r}_{ui}$ and the true rating $r_{ui}$, 

\begin{equation}
    \text{RMSE} = \sqrt{\frac{1}{|\mathcal{T}|} \sum_{(u,i) \in \mathcal{T}} (r_{ui} - \hat{r}_{ui})^2}
\end{equation}

where $\mathcal{T}$ denotes the set of user-item pairs in the test set \citep{salako2024fish}.

\paragraph{Precision@K}
While RMSE measures rating accuracy, it does not necessarily correlate with the ability to recommend relevant items. We evaluate ranking performance by generating top-$K$ recommendations for a subset of test users. For evaluation, we consider an item to be relevant if the ground truth rating is $\ge 3.5$.

\begin{equation}
\begin{split}
\text{Precision@K} &= \frac{1}{|\mathcal{U}_{\text{test}}|} 
    \sum_{u \in \mathcal{U}_{\text{test}}} 
    \frac{|\{ i \in \text{TopK}(u) : r_{ui} \geq 3.5 \}|}{K}
\end{split}
\end{equation}

This metric quantifies the proportion of recommended items that the user actually liked.

\paragraph{Recall@K}
Recall measures the ability of the system to find all relevant items in the test set. It is given as:

\begin{equation}
\begin{split}
\text{Recall@K} &= \frac{1}{|\mathcal{U}_{\text{test}}|} 
\sum_{u \in \mathcal{U}_{\text{test}}} 
\frac{|\{i \in \text{TopK}(u) : r_{ui} \geq 3.5\}|}
     {|\{i \in \mathcal{T}_u : r_{ui} \geq 3.5\}|}
\end{split}
\end{equation}

where $\mathcal{T}_u$ is the set of items rated by user $u$ in the test set. In our experiments, we generate predictions for all unobserved items and test items, masking training items to prevent data leakage.

\subsection{Hyperparameter Tuning}
We conduct an exhaustive Grid Search to identify the optimal configuration for training. We explore the space shown in Table \ref{tab:hyperparams}. We trained separate models for every combination of $k, \lambda, \tau$ for 20 epochs. To maintain computational feasibility during the search, Precision and Recall metrics are evaluated on a random sample of 3,000 users, while RMSE is calculated on the full test set. We then analyze the trade-off between model complexity and generalization performance to select the final model parameters. All models were trained using complementary compute provided on Google Colab.

\begin{table}[tb!]
\centering
\caption{Hyperparameter search space explored for model
training}
\label{tab:hyperparams}
\begin{tabular}{l c c}
\toprule
\textbf{Parameter}              & \textbf{Symbol} & \textbf{Values}                  \\
\midrule
Latent dimensions               & $k$             & 2, 10, 50, 100            \\
Rating error weight             & $\lambda$       & 0.1, 0.5                   \\
L2 regularization     & $\tau$          & 0.05, 0.1, 0.25           \\
\bottomrule
\end{tabular}
\end{table}

\section{Results and Conclusion}
\label{result}

We conducted an extensive grid search to evaluate the performance of the MF model across distinct hyperparameters. Table \ref{tab:results} shows the outcome of the search and highlights the delicate balance between model capacity and generalization. We observed that increasing the latent dimension significantly reduced the Training RMSE, reaching a low of $0.415$ at $k=100$. However, this improvement came at the cost of severe overfitting, as evidenced by the Test RMSE rising to over $1.0$ in high-dimensional configurations. Conversely, the low-capacity model, $k=2$, underfitted the data, yielding poor ranking metrics across the board.

\begin{figure*}[tb!]
\centering
\includegraphics[width=0.98\textwidth]{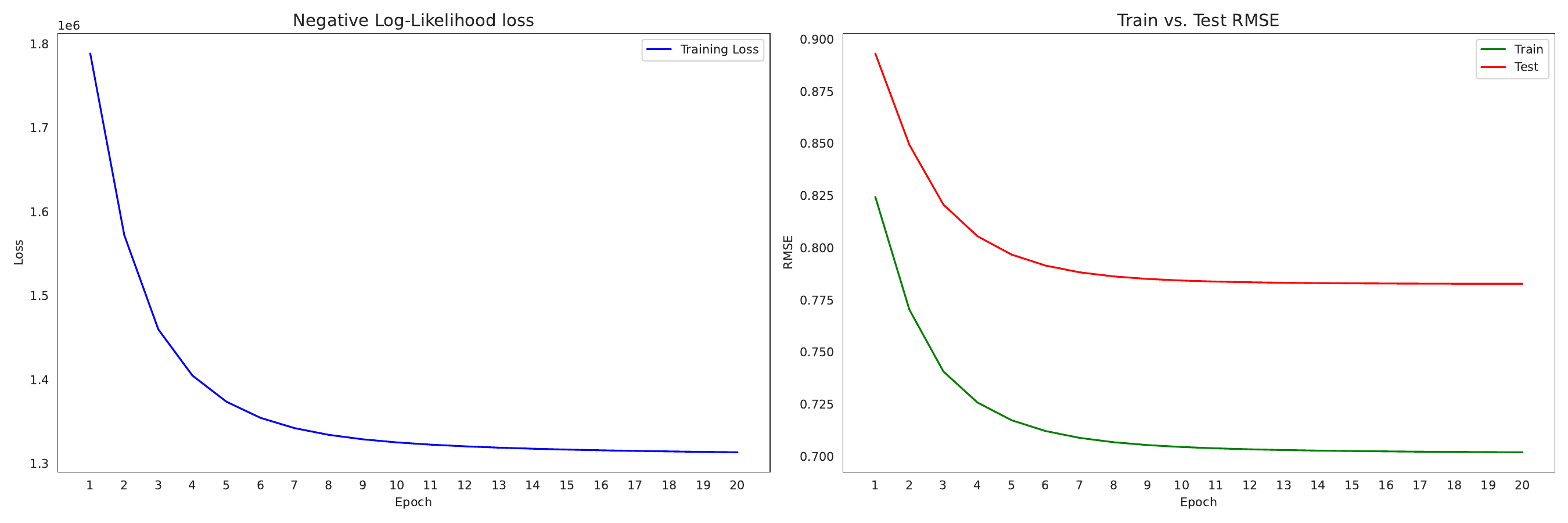}
\caption{Training dynamics of the best performing model (k=10,
$\lambda$=0.1, $\tau$=0.25).}
\label{fig:best_model_history}
\end{figure*}

\begin{table}[tb!]
\centering
\caption{Hyperparameter grid search results for the MF model,
including Training RMSE and Test RMSE}
\label{tab:results}
\resizebox{\columnwidth}{!}{
\begin{tabular}{@{}ccccccc@{}}
\toprule
\textbf{k} & \textbf{$\lambda$} & \textbf{$\tau$} & \textbf{Precision@k} & \textbf{Recall@k} & \textbf{Train RMSE} & \textbf{Test RMSE} \\ \midrule
2 & 0.1 & 0.05 & 0.0000 & 0.0000 & 0.7873 & 0.8157 \\
2 & 0.1 & 0.10 & 0.0000 & 0.0000 & 0.7865 & 0.8139 \\
2 & 0.1 & 0.25 & 0.0082 & 0.0015 & 0.7884 & 0.8135 \\
2 & 0.5 & 0.05 & 0.0000 & 0.0000 & 0.7851 & 0.8158 \\
2 & 0.5 & 0.10 & 0.0000 & 0.0000 & 0.7855 & 0.8154 \\
2 & 0.5 & 0.25 & 0.0000 & 0.0000 & 0.7858 & 0.8145 \\ \midrule
10 & 0.1 & 0.05 & 0.0028 & 0.0018 & 0.6959 & 0.7924 \\
10 & 0.1 & 0.10 & 0.0159 & 0.0098 & 0.6980 & 0.7877 \\
\textbf{10} & \textbf{0.1} & \textbf{0.25} & \textbf{0.0430} & \textbf{0.0257} & \textbf{0.7020} & \textbf{0.7828} \\
10 & 0.5 & 0.05 & 0.0000 & 0.0000 & 0.6940 & 0.8047 \\
10 & 0.5 & 0.10 & 0.0000 & 0.0000 & 0.6948 & 0.7995 \\
10 & 0.5 & 0.25 & 0.0023 & 0.0013 & 0.6957 & 0.7919 \\ \midrule
50 & 0.1 & 0.05 & 0.0043 & 0.0085 & 0.5302 & 0.9041 \\
50 & 0.1 & 0.10 & 0.0119 & 0.0284 & 0.5342 & 0.8834 \\
50 & 0.1 & 0.25 & 0.0258 & 0.0713 & 0.5446 & 0.8574 \\
50 & 0.5 & 0.05 & 0.0001 & 0.0002 & 0.5278 & 0.9645 \\
50 & 0.5 & 0.10 & 0.0006 & 0.0009 & 0.5282 & 0.9360 \\
50 & 0.5 & 0.25 & 0.0043 & 0.0085 & 0.5303 & 0.9040 \\ \midrule
100 & 0.1 & 0.05 & 0.0043 & 0.0211 & 0.4176 & 0.9967 \\
100 & 0.1 & 0.10 & 0.0101 & 0.0577 & 0.4225 & 0.9667 \\
100 & 0.1 & 0.25 & 0.0199 & 0.1149 & 0.4361 & 0.9304 \\
100 & 0.5 & 0.05 & 0.0003 & 0.0008 & 0.4157 & 1.0733 \\
100 & 0.5 & 0.10 & 0.0009 & 0.0031 & 0.4157 & 1.0377 \\
100 & 0.5 & 0.25 & 0.0043 & 0.0217 & 0.4176 & 0.9952 \\
\bottomrule
\end{tabular}
}
\end{table}

The optimal configuration was found at $k=10$, $\lambda=0.1$, and $\tau=0.25$. This setting achieved the lowest Test RMSE of $0.7828$ and the highest ranking performance with a Precision@10 of $0.043$. Notably, stronger regularization, $\tau=0.25$, consistently improved performance, suggesting that the sparsity of the MovieLens 32M dataset requires significant constraints on the magnitude of latent vectors. We visualize the learning dynamics of this optimal configuration in Figure \ref{fig:best_model_history}. The left panel depicts the monotonic decrease of the regularized negative log-likelihood, while the right panel reveals that the model converges rapidly, with the Test RMSE stabilizing around epoch 15. Unlike higher dimensional models, where the Test RMSE begins to diverge from the Training RMSE significantly, the gap here remains controlled.

\subsection{Movie Trait Vectors}

\begin{figure*}[tb!]
    \centering
    \includegraphics[width=0.95\linewidth]{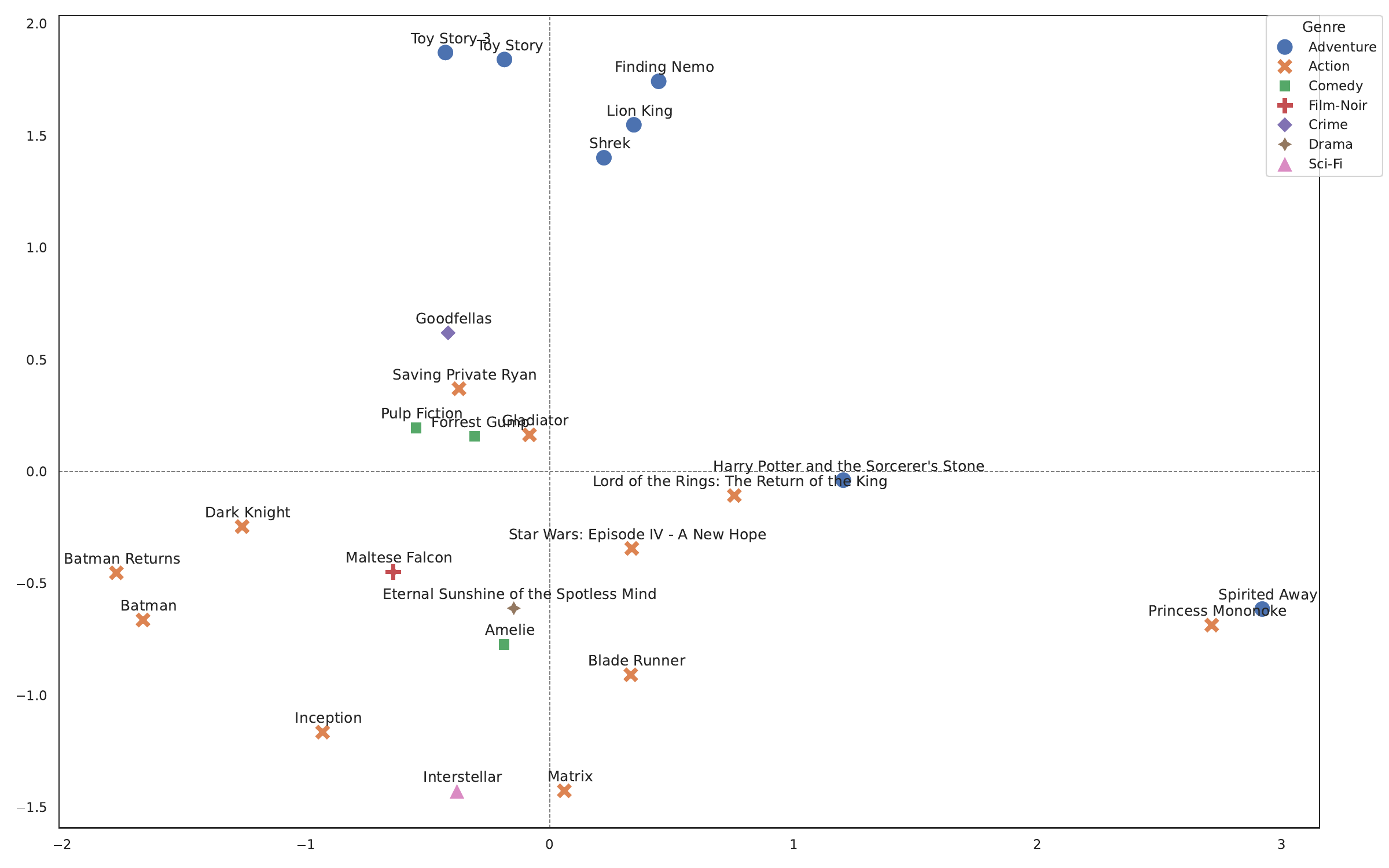}
    \caption{Visualization of latent trait vectors for selected movies, projected into 2D space using PCA.}
    \label{fig:vectors}
\end{figure*}

To interpret the latent space learned by the model, we extracted the item trait vectors $\mathbf{v}_i$ for a selection of movies representing distinct genres. Because the optimal model utilizes a latent dimension $k$ greater than 2, we applied Principal Component Analysis (PCA) to project these high-dimensional vectors into a 2D space for visualization. As illustrated in Figure \ref{fig:vectors}, the model successfully encodes semantic similarity into geometric proximity. Movies sharing genres or thematic elements naturally cluster together. For instance, animations like \textit{The Lion King} and \textit{Shrek} occupy a distinct region of the vector space, separated from crime dramas like \textit{Goodfellas}. This confirms that the ALS algorithm recovered meaningful latent structures solely from interaction data, without access to metadata tags.

\subsection{Movie Recommendation}

To assess the utility of the model, we synthesized a dummy user profile exhibiting a strong affinity for animation. We assigned 5.0 star ratings to three specific items: \textit{Toy Story (1995)}, \textit{Toy Story 3 (2010)}, and \textit{The Lion King (1994)}. Rather than retraining the full model, we calculated the user's latent vector using a single iteration of the ALS user step, keeping all item parameters fixed. We generated recommendations using a modified scoring function that introduces a hyperparameter $\alpha$ to weight the influence of global popularity biases against personalized affinity:

\begin{equation}
    \text{Score} = \alpha (\mu + b_i) + \mathbf{u}_u^T \mathbf{v}_i
\end{equation}

where $\mu + b_i$ represents the baseline popularity of an item, while $\mathbf{u}_u^T \mathbf{v}_i$ captures the specific interaction between user taste and item traits. To ensure robustness, we filtered out items with fewer than 100 total ratings.

Table \ref{tab:recommendations} presents the top recommendations generated under different $\alpha$ regimes. The results demonstrate the critical role of bias weighting in recommendation diversity:

\begin{itemize}
    \item \textbf{High Personalization ($\alpha=0.05$):} The model prioritizes the dot product term. As seen in the first column of Table \ref{tab:recommendations}, the system successfully retrieves direct sequels (\textit{Toy Story 2}, \textit{Toy Story 4}) and thematically similar animations (\textit{Inside Out}, \textit{Finding Dory}).
    \item \textbf{High Popularity ($\alpha=1.0$):} The global bias term dominates the score. The recommendations shift entirely toward universally acclaimed content. While these items are of high quality, they lack personalization based on the user's specific input history.
    \item \textbf{$\alpha=0.5$:} This setting offers a middle ground, recommending high-rated items that still retain some relevance to the user's latent preferences.
\end{itemize}

\begin{table*}[th!]
\centering
\caption{Top recommendations for dummy user.}
\label{tab:recommendations}
\small
\begin{tabular}{c l l l}
\toprule
Rank & $\alpha=0.05$ & $\alpha=0.5$ & $\alpha=1.0$ \\
\midrule
1  & Irreversible (2002)                    & Planet Earth II (2016)            & Planet Earth II (2016)            \\
2  & Toy Story 2 (1999)                     & Planet Earth (2006)               & Planet Earth (2006)               \\
3  & Toy Story 4 (2019)                     & Over the Garden Wall (2014)       & Band of Brothers (2001)           \\
4  & Inside Out (2015)                      & Death Note (2006)                 & The Blue Planet (2001)            \\
5  & Feast (2014)                           & Powers of Ten (1977)              & Blue Planet II (2017)             \\
6  & Finding Dory (2016)                    & Cosmos: A Spacetime Odyssey (2014)& Twelve Angry Men (1957)           \\
7  & Harvie Krumpet (2003)                  & HyperNormalisation (2016)         & Powers of Ten (1977)              \\
8  & Babylon (2022)                         & Cosmos (1980)                     & Cosmos (1980)                     \\
9  & Winnie Pooh (1969)                     & Band of Brothers (2001)           & Alone in the Wilderness (2004)    \\
10 & Toy Story That Time Forgot (2014)      & The Meeting Place (1979)          & Over the Garden Wall (2014)       \\
\bottomrule
\end{tabular}
\end{table*}

This experiment confirms that the learned latent vectors capture meaningful semantic relationships, and that $\alpha$ serves as an effective tunable parameter to manage the trade-off between serendipitous personalization and safe, popular recommendations.

\section{Conclusion}

In this work, we investigated the latent geometry of user preferences within the MovieLens 32M dataset, employing a regularized Matrix Factorization framework optimized via Alternating Least Squares. Our hyperparameter analysis reveals that increasing model capacity does not monotonically improve recommendation quality in sparse environments. While higher dimensional latent spaces minimize training error, they suffer significantly from overfitting. A model with $k=10$ dimensions and robust regularization emerged as the optimal configuration. Geometric analysis of the learned embeddings demonstrates that the model successfully encodes semantic information with the emergence of distinct genre clusters within the vector space, confirming that the interaction matrix contains sufficient signal to reconstruct thematic relationships. Future work may extend this analysis to temporal dynamics, exploring how these latent geometries evolve as user tastes shift over time.

\bibliography{ref}

@article{koren2009matrix,
  title={Matrix factorization techniques for recommender systems},
  author={Koren, Yehuda and Bell, Robert and Volinsky, Chris},
  journal={Computer},
  volume={42},
  number={8},
  pages={30--37},
  year={2009},
  publisher={IEEE}
}

@inproceedings{sarwar2001item,
  title={Item-based collaborative filtering recommendation algorithms},
  author={Sarwar, Badrul and Karypis, George and Konstan, Joseph and Riedl, John},
  booktitle={Proceedings of the 10th international conference on World Wide Web},
  pages={285--295},
  year={2001}
}

@inproceedings{koenigstein2012xbox,
  title={The Xbox recommender system},
  author={Koenigstein, Noam and Nice, Nir and Paquet, Ulrich and Schleyen, Nir},
  booktitle={Proceedings of the sixth ACM conference on Recommender systems},
  pages={281--284},
  year={2012}
}

@inproceedings{salakhutdinov2007probabilistic,
  title={Probabilistic matrix factorization},
  author={Salakhutdinov, Ruslan and Mnih, Andriy},
  booktitle={Advances in neural information processing systems},
  volume={20},
  year={2007}
}

@inproceedings{zhou2008large,
  title={Large-scale parallel collaborative filtering for the netflix prize},
  author={Zhou, Yunhong and Wilkinson, Dennis and Schreiber, Robert and Pan, Rong},
  booktitle={International conference on algorithmic applications in management},
  pages={337--348},
  year={2008},
  organization={Springer}
}

@article{salako2024fish,
  title={Fish-NET: Advancing aquaculture management through AI-enhanced fish monitoring and tracking},
  author={Salako, Joshua and Ojo, Foluso and Awe, Olushina Olawale},
  journal={AGRIS on-line Papers in Economics and Informatics},
  volume={16},
  number={2},
  pages={121--131},
  year={2024}
}
\bibliographystyle{icml2023}

\newpage
\appendix
\onecolumn
\section{Movie Recommender Application}

\begin{figure*}[h!]
    \centering
    
    \begin{subfigure}[b]{0.8\textwidth}
        \centering
        \includegraphics[width=\linewidth]{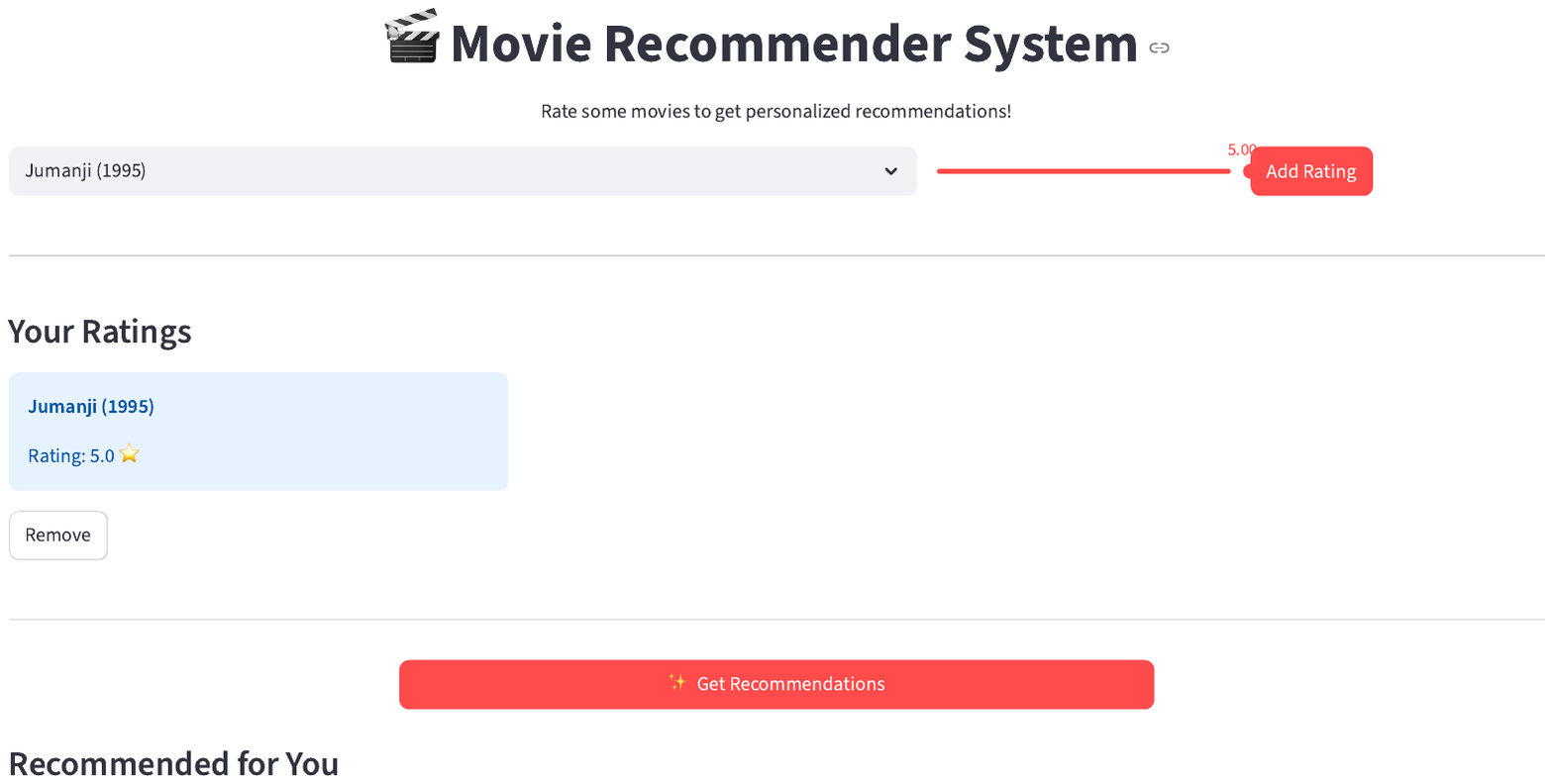}
    \end{subfigure}    
    \begin{subfigure}[b]{0.8\textwidth}
        \centering
        \includegraphics[width=\linewidth]{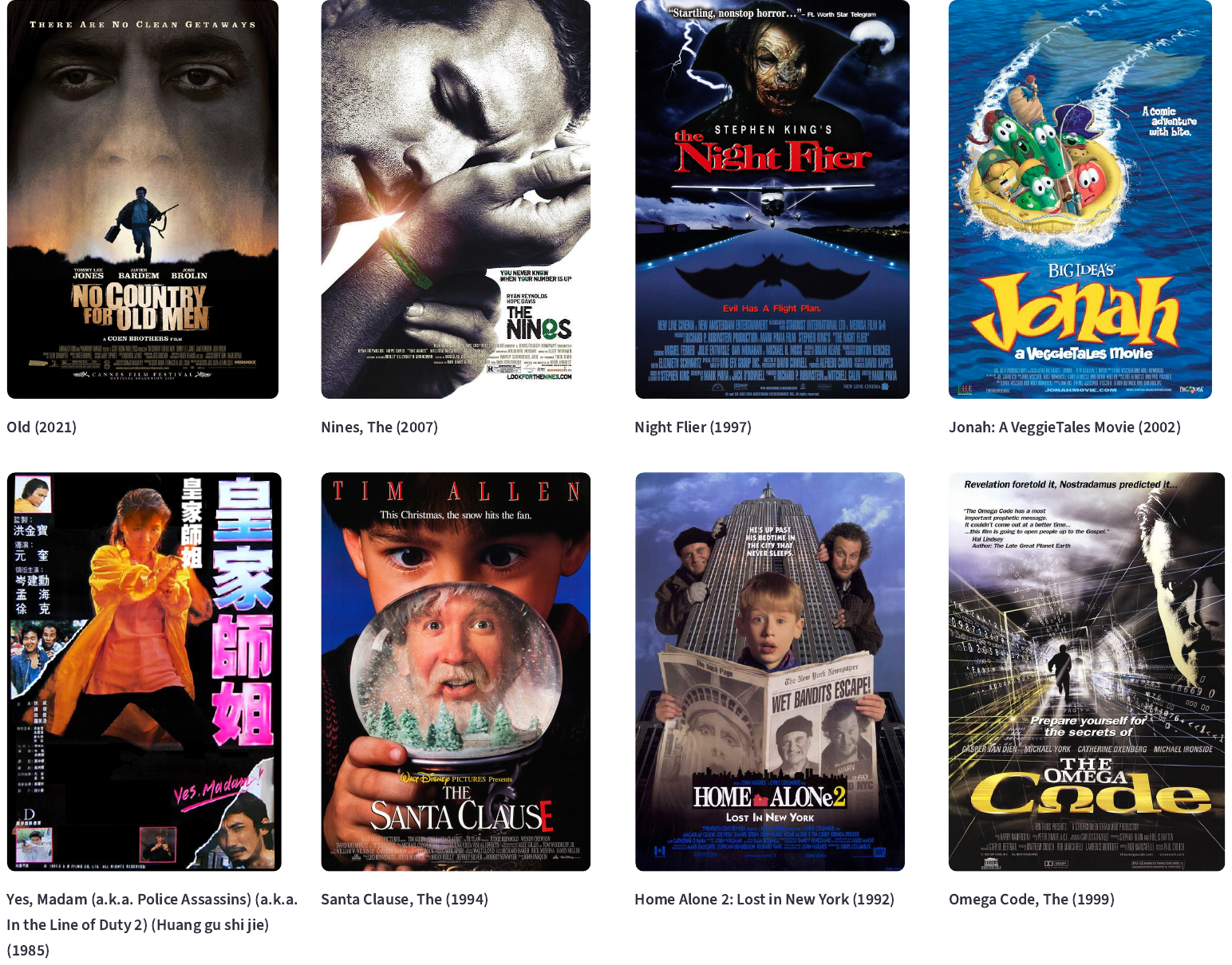}
    \end{subfigure}
    
    \caption{Recommender application interface.}
    \label{fig:application}
\end{figure*}

The best performing model was deployed as a Streamlit application. Figure \ref{fig:application} illustrates the application interface, where users can search the full movie catalog, assign ratings to movies they have watched, and receive personalized recommendations inferred from their expressed preferences. Live demo can be found here: \emph{https://als-recommender.streamlit.app/}

\end{document}